\documentclass[runningheads]{llncs}

\usepackage{eccv}

\usepackage{eccvabbrv}

\usepackage{graphicx}
\usepackage{booktabs}
\usepackage[misc]{ifsym}
\usepackage{array}
\usepackage{soul}
\usepackage{color}
\usepackage{xcolor}
\usepackage{bbm}
\usepackage{multirow}
\usepackage[ruled,vlined]{algorithm2e} 
\usepackage{setspace}
\usepackage{amsmath}
\usepackage{tikz}
\usepackage{caption} 

\definecolor{mygray}{gray}{.92}
\makeatletter
\def\blfootnote{\xdef\@thefnmark{}\@footnotetext}
\makeatother

\usepackage[accsupp]{axessibility}  

\usepackage{hyperref}

\usepackage{orcidlink}

\begin{document}

\title{PosFormer: Recognizing Complex\\ Handwritten Mathematical Expression with Position Forest Transformer} 

\titlerunning{Position Forest Transformer for HMER}

\author{Tongkun Guan\inst{1}\textsuperscript{$*$}\orcidlink{0000-0003-3346-8315} \and
Chengyu Lin\inst{2}\textsuperscript{$*$}\orcidlink{0009-0000-2740-2315} \and
Wei Shen\inst{1(\textrm{\Letter})} \and
Xiaokang Yang\inst{1}
}

\authorrunning{T.~Guan et al.}

\institute{MoE Key Lab of Artificial Intelligence, AI Institute, Shanghai Jiao Tong University \email{\{gtk0615,wei.shen\}@sjtu.edu.cn} \and
Paris Elite Institute of Technology, Shanghai Jiao Tong University
\email{lacayqwq@sjtu.edu.cn}\\
\url{https://github.com/SJTU-DeepVisionLab/PosFormer}
}

\maketitle
\blfootnote{\noindent$^{*}$Equal contribution.\\  \textsuperscript{\Letter}Corresponding author.}

\begin{abstract}
 Handwritten Mathematical Expression Recognition (HMER) has wide applications in human-machine interaction scenarios, such as digitized education and automated offices. Recently, sequence-based models with encoder-decoder architectures have been commonly adopted to address this task by directly predicting LaTeX sequences of expression images. However, these methods only implicitly learn the syntax rules provided by LaTeX, which may fail to describe the position and hierarchical relationship between symbols due to complex structural relations and diverse handwriting styles. To overcome this challenge, we propose a \underline{pos}ition \underline{for}est transfor\underline{mer} (PosFormer) for HMER, which jointly optimizes two tasks: expression recognition and position recognition, to explicitly enable position-aware symbol feature representation learning. Specifically, we first design a position forest that models the mathematical expression as a forest structure and parses the relative position relationships between symbols. Without requiring extra annotations, each symbol is assigned a position identifier in the forest to denote its relative spatial position. Second, we propose an implicit attention correction module to accurately capture attention for HMER in the sequence-based decoder architecture. Extensive experiments validate the superiority of PosFormer, which consistently outperforms the state-of-the-art methods 2.03\%/1.22\%/2.00\%, 1.83\%, and 4.62\% gains on the single-line CROHME 2014/2016/2019, multi-line M$^{2}$E, and complex MNE datasets, respectively, with no additional latency or computational cost.
\end{abstract}

\section{Introduction}
\label{sec:intro}

Handwritten mathematical expressions, serving as a nexus between the language and symbols, are common in fields including mathematics, physics, and chemistry. The corresponding task, known as Handwritten Mathematical Expression Recognition (HMER), aims to accurately convert expression images into LaTeX sequences. This task has wide applications in human-machine interaction scenarios like online education, manuscript digitization, and automatic scoring. 
However, recognizing these expressions poses two distinct challenges: 1) the complexity of inter-symbol relationships\cite{anderson1967syntax}, which makes the model struggle to generate appropriate structural symbols regulated by LaTeX; and 2) the diversity of handwriting inputs, including variations in scale and style.

Existing methods introduce or develop advanced components for symbol recognition and/or structural analysis to enhance recognition capabilities. They can be summarized into two branches: tree-based methods\cite{yuan2022syntax,li2024tree} and sequence-based methods \cite{zhang2017watch,zhao2022comer,yang2023read}.
Specifically, following the syntax rules of LaTeX, tree-based methods model each mathematical expression as a tree structure\cite{zhang2017tree}, and then output sequences about the complete triple tuples (parent, child, parent-child relationships) of the syntax tree and decode them into a LaTeX sequence.
These methods exhibit lower accuracy and poor generalization, limited by the insufficient diversity of tree structure across expressions.
Sequence-based methods, which model HMER as an end-to-end image-to-sequence task\cite{shi2016end}, have gained increasing attention. They view the mathematical expression as a LaTeX sequence and employ an attention-based encoder-decoder architecture to predict each symbol in an autoregressive manner. However, these methods only implicitly learn the structure relationships between symbols and fall short in processing complex and nested mathematical expressions.

To address this issue, we propose a Position Forest Transformer (PosFormer), which explicitly models structure relationships between symbols in the attention-based encoder-decoder models to enable complex mathematical expression recognition. 
Specifically, we encode the LaTeX mathematical expression sequence as a position forest structure, where each symbol is assigned a position identifier to denote its relative spatial position in a two-dimensional image. 
Leveraging this position forest coding, we parse the nested levels and relative positions of each symbol in the forest to assist position-aware symbol-level feature representation learning in complex and nested mathematical expressions.
Additionally, we introduce an implicit attention correction module in the attention-based decoder architecture to enhance attention precision. By adaptively incorporating zero attention as a refinement term, we refine attention weights with past alignment information, leading to more fine-grained feature representations.

It is noteworthy that our proposed PosFormer method just requires original HMER annotations (LaTeX sequences) without extra labelling work, and incurs no additional latency or computational cost during the inference stage. 
In general, our contributions are summarized as follows:

\begin{itemize}
      \setlength{\itemsep}{-0pt}
      \setlength{\parsep}{-0pt}
      \setlength{\parskip}{0pt}
    \item We introduce a new method for HMER, PosFormer, which models the mathematical expression as a position forest structure and explicitly parses the relative position relationships between symbols to significantly improve the end-to-end image-to-sequence recognition capability.
    \item PosFormer achieves state-of-the-art (SOTA) performance on the publicly available single-line and multi-line benchmarks, with 2.03\%/1.22\%/2.00\% and 1.83\% gains on the CROHME 2014\cite{mouchere2014icfhr}/2016\cite{mouchere2016icfhr2016}/2019\cite{mahdavi2019icdar} and M$^{2}$E\cite{yang2023read} datasets, respectively.
    When recognizing the complex MNE dataset, PosFormer further exhibits a remarkable average gain of 4.62\%.
\end{itemize}

\section{Related Work}
Handwritten Mathematical Expression Recognition (HMER) has attracted considerable attention, due to the challenge of handling complicated text images with a variety of handwriting styles, structure complexity, \etc. Traditional methods\cite{chan1998elastic,kosmala1999line,hu2011hmm,winkler1996hmm,vuong2010towards,keshari2007hybrid} show low accuracy and typically involve a two-step pipeline: recognizing individual symbols and subsequently correcting them guided by grammatical rules\cite{chan2000mathematical}. 
 Recently, with the development of deep learning~\cite{guan2022industrial,guan2023bridging}, two mainstream methods have been developed to improve recognition performance: tree-based methods \cite{zhang2017tree,9147045,wu2021graph,yuan2022syntax,TSDNet,zhang2020tree,wu2022tdv2,li2024tree} and sequence-based methods\cite{SIGA,zhang2017watch,deng2017image,zhang2018multi,huang2017densely,li2022counting,zhao2021handwritten,zhao2022comer,yang2023read,le2020recognizing,le2019pattern,wang2019multi,zhang2018track,truong2020improvement,wu2020handwritten}.

\noindent \textbf{Tree-based Methods.}
Tree-based methods view the mathematical expression as a tree structure and develop tree-structured decoders to model the hierarchical relationship based on the corresponding syntax rules. 
Specifically, early tree-based methods are developed to recognize online datasets, which contain writing trajectory information. BLSTM\cite{zhang2017tree} proposes a pioneering work that encodes each mathematical expression into a tree by modeling symbols as nodes and relationships between symbols as edges. Utilizing the uniqueness of strokes, SRD\cite{9147045} proposes a sequential relationship decoder, which improves the recognition performance of tree structures.
Recently, with the development of tree-based methods, some approaches have expanded to more challenging offline HMER tasks. 
TSDNet\cite{TSDNet} employs a Transformer-based multi-task learning to jointly model and predict the node attributes, edge attributes, and node connectivities, capturing the structural correlations among tree nodes. SAN\cite{yuan2022syntax} is proposed to enhance syntax awareness by introducing grammatical constraints.

\noindent \textbf{Sequence-based Methods.}
Considering image features as sequences, these methods employ an autoregressive decoder framework for HMER. Specifically, a pioneering method introduced by Deng \emph{et al.}\cite{deng2017image} converts images into LaTeX sequences. They employ an RNN decoder to sequentially recognize symbols, with contextual priors provided by previously recognized symbols. WAP\cite{zhang2017watch} utilizes a CNN for visual feature extraction and a GRU for decoding LaTeX sequences. By incorporating a coverage attention mechanism into the GRU, the model alleviates the over-translation problem.
Building on this, DWAP\cite{zhang2018multi} incorporates a multi-scale DenseNet\cite{huang2017densely} encoder to strengthen feature extraction and facilitate gradient propagation. CAN\cite{li2022counting} introduces a weakly supervised counting task as an auxiliary branch to assist the recognition. 
Additionally, BTTR\cite{zhao2021handwritten} is the first to apply a transformer structure for HMER, which utilizes bidirectional decoding to mitigate error accumulation. Inspired by the coverage information mechanism in RNNs, CoMER\cite{zhao2022comer} further develops an attention refinement module to solve the coverage problem. LAST\cite{yang2023read} further introduces a line-aware semi-autoregressive transformer to focus on multi-line recognizing tasks.

Differing from these methods, we model the mathematical expression as a position forest structure and parse the nested levels and relative positions between symbols to explicitly enable position-aware symbol-level feature representation learning in sequence-based encoder-decoder models. 
Accordingly, our PosFormer with position awareness can effectively generate ``\texttt{\^{}}'',``\texttt{\_}'',``\texttt{\{}'' and ``\texttt{\}}'' to obtain accurate LaTeX sequences, especially when recognizing complex expressions.

\section{Methodology}
\begin{figure}[t]
  \centering
  \includegraphics[width=1.0\linewidth]{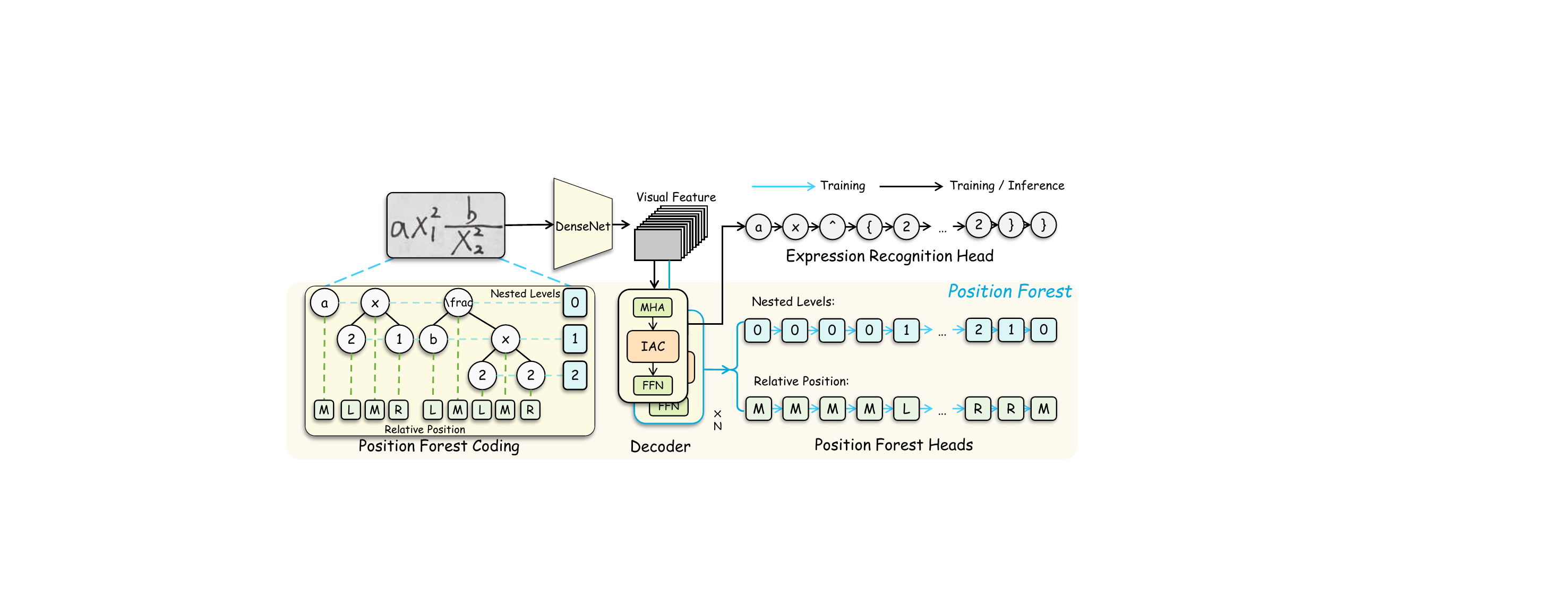}
  \caption{Overall structure of our proposed PosFormer, which jointly optimizes two tasks: expression recognition and position recognition. The former employs parallel linear prediction for symbol recognition; the latter encodes the LaTeX sequence as a position forest structure and decodes the nested levels and relative positions of each symbol to assist in position-aware symbol-level feature representation learning.}
  \label{fig:PosFormer}
\end{figure}
\subsection{Overview}
\noindent \textbf{Problem definition:} Given a handwritten mathematical expression image $\mathbf{X}\in \mathbb{R}^{H \times W}$, we aim to interpret the mathematical expression and get the corresponding sequence $\mathcal{Y}_{c}=\big\{y_{c}^{(t)}|y_{c}^{(t)} \in \{$``\texttt{a}'', ``\texttt{b}'', $\cdots$, ``\texttt{\_}''$\}\big\}_{t=1}^{T}$. H, W, and T are the image height, image width, and sequence length, respectively.
For simplification, these notations are the same on all images within the scope of this paper.

\noindent \textbf{Network pipeline:} We employ the sequence-based encoder-decoder method, CoMER\cite{zhao2022comer}, as our baseline model. As shown in Figure \ref{fig:PosFormer}, the Position Forest Transformer (PosFormer) consists of a DenseNet backbone, our position forest, and an expression recognition head. 
\noindent 1) \texttt{Training:} Initially, the backbone extracts 2D visual features from the input image. These features are subsequently fed into the attention-based transformer decoder to obtain discriminative symbol features. A parallel linear head is then deployed to recognize the LaTeX expression. Together with expression recognition, a position forest is introduced for joint optimization to facilitate the learning of position-aware symbol-level feature representations. Specifically, this process begins by encoding the sequence $\mathcal{Y}_{c}=\{y_{c}^{(t)}\}_{t=1}^{T}$ of mathematical expression into an identifier set $\mathcal{I}=\{\mathbf{I}_{t}\}_{t=1}^{T}$, as outlined in Algorithm \ref{algorithm0}. Each identifier denotes a string that represents its position information. Leveraging this encoding, two position forest heads are then employed to parse their nested levels and relative positions in the forest, details of which will be introduced later.
\noindent 2) \texttt{Inference:} The input image is sequentially passed through the backbone, the decoder, and the expression recognition head to predict the LaTeX sequence. Note that the position forest coding and the position forest heads are removed during inference, which brings no additional latency or computational cost.

\subsection{Position Forest Transformer} 
In this section, we will introduce the detail of our position forest transformer (PosFormer), which explicitly enhances the ability to perceive relative position relationships for recognizing complex and nested handwritten mathematical expressions. Specifically, we will introduce the position forest and the implicit attention correction module.
\begin{figure}[t]
  \centering
  \includegraphics[width=0.9\linewidth]{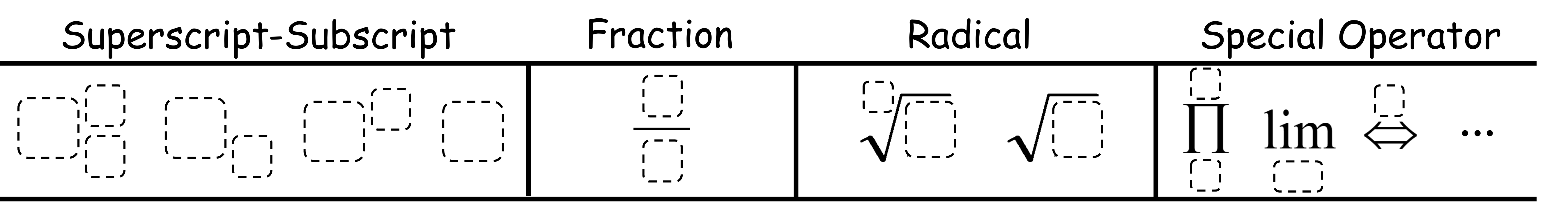}
  \caption{Four substructure types of mathematical expressions, including superscript-subscript, fraction, radical, and special operator structures.}
  \label{fig:substructuretype}
\end{figure}
\begin{figure}[t]
  \centering
  \includegraphics[width=0.9\linewidth]{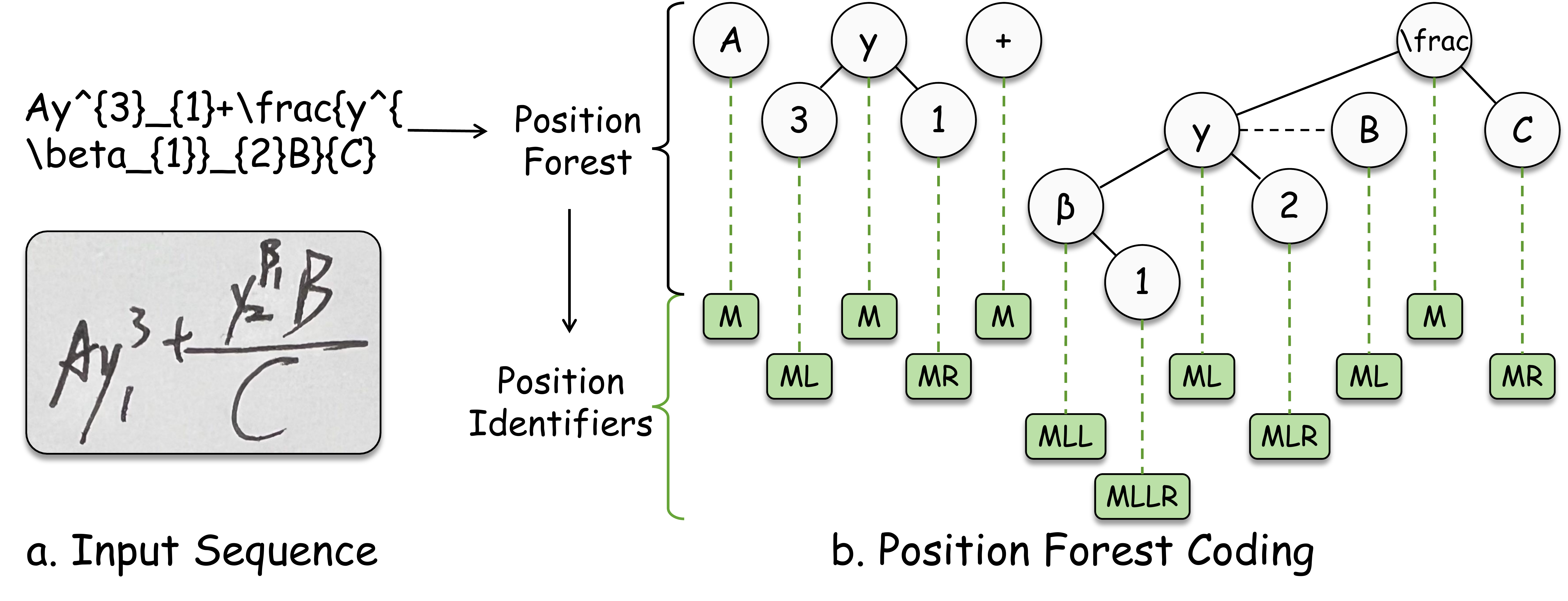}
  \caption{Illustration of the position forest coding process, which can be simply described as sequence $\rightarrow$ substructure $\rightarrow$ tree $\rightarrow$ position forest. Specifically, we encode each symbol as a position identifier to denote its relative spatial position (\eg, ``MLLR'').}
  \label{fig:PCT}
\end{figure}

\begin{figure}[!t]
\begin{algorithm}[H]
\DontPrintSemicolon
\setstretch{0.9}
\fontsize{8pt}{9pt}\selectfont
\caption{\textbf{Position Forest Coding}}
\label{algorithm0}
\SetKwProg{Fn}{Function}{\string:}{}
\SetKwFunction{encoding}{Substruct2Identifier}
\SetKwFunction{recursion}{Sequence2Substruct}
\SetKwFunction{FindEnd}{Find\_End}
\nl {\bf Input}: A sequence $\mathcal{Y}_{c}=\{y_{c}^{(t)}\}_{t=1}^{T}$ of length T\;
\nl Define the structure type set $\{\theta_{k}\}_{k=1}^{4} = \{$ ``\texttt{\^{}}'', ``\texttt{\_}'', ``\texttt{$\backslash sqrt$}'', ``\texttt{$\backslash frac$}'' $\}$ \;
\nl Define ``M''/``L''/``R'' as the identifiers for the root/left/right node of a tree \;
\nl Define an identifier set $\mathcal{I}=\{\mathbf{I}_{t}\}_{t=1}^{T}$ for encoding the positional relationship of all symbols, and initialize all elements to ``M''\;
\nl \Fn{\encoding{l, r, $y_{c}^{(l)}$}}{
    \If{$y_{c}^{(l)} \in \{\theta_{1}\}$}{
            \nl $\mathbf{I}_{t} \text{ += ``L''} \quad \forall t\in [l,r]$\;
        }
    \If{$y_{c}^{(l)} \in \{\theta_{2}, \theta_{3}\}$}{
            \nl $\mathbf{I}_{t} \text{ += ``R''} \quad \forall t\in [l,r]$\;
        }
    \If{$y_{c}^{(l)} \in \{\theta_{4}\}$}{
            \nl $r_{1}, r_{2} = r$\;
            \nl $\mathbf{I}_{t} \text{ += ``L''} \quad \forall t\in [l,r_{1}]$; \,
            $\mathbf{I}_{t} \text{ += ``R''} \quad \forall t\in (r_{1},r_{2}]$\;
        }
}
\nl \Fn{\recursion{$l$, $r$}}
{
\nl \While{$l < r$}{
    \If{$y_{c}^{(l)} \in \{\theta_{k}\}_{k=1}^{3}$ }{
     \nl Search its corresponding substructure to get the position index $i_e\in \mathbb{R}^{T}$ of last symbol ``$\}$'' in the substructure\;
     \nl \encoding{l, $i_e$, $y_{c}^{(l)}$}\tcp*{Position encoding}
    \nl \recursion{l, $i_e$}\tcp*{Recursive search}
    \nl $l \gets i_e + 1$\;
    }
    \ElseIf{$y_{c}^{(l)} \in \{\theta_{4}\}$}{
    \nl Search its corresponding substructure (``$\backslash frac\{\}\{\}$'') to get two position indexes $i_{e1},i_{e2}\in \mathbb{R}^{T}$ of symbol ``$\}$''\;
        \nl \encoding{$l$, $(i_{e1},i_{e2})$, $y_{c}^{(l)}$}\tcp*{Position encoding}
        \nl \recursion{$l$, $i_{e2}$}\tcp*{Recursive search}
        \nl $l \gets i_{e2} + 1$\;
    }
    \Else{
        \nl $l \gets l + 1$\;
    }
}
}
\nl \recursion{0, T}\;
\nl {\bf Return} $\mathcal{I}=\{\mathbf{I}_{t}\}_{t=1}^{T}$\tcp*{$\mathbf{I}_{t}$ denotes a string, consisting of M, L, and R}
\end{algorithm}
\end{figure}

\noindent \textbf{Position Forest}
Given a handwritten expression, we aim to encode the corresponding LaTeX sequence to model structure relationships between symbols, and parse them to assist the sequence-based encoder-decoder models.

\noindent 1) \texttt{Position Forest Coding:} According to the syntax rules in LaTeX,  expressions can be easily divided into several substructures as depicted in Figure \ref{fig:substructuretype}, including superscript-subscript structures, fraction structures, radical structures, and special operator structures (\eg, Product, Limit, and Equivalent). 

These substructures exhibit either independent or nested relationships. Within each substructure, the relative position relationships of symbols are categorized into three types: ``upper'', ``lower'', and ``middle'', based on their spatial positions in an image.
Leveraging this prior knowledge, we model the LaTeX mathematical expression as a position forest structure. The construction follows three rules: 

 1) These substructures are encoded in a left-to-right order;
 
2) Each substructure is encoded into a tree based on the relative positions between symbols, with its main body as the root node, its upper part as the left node, and its lower part as the right node;
 
 3) According to the substructure relationships, these encoded trees are arranged in series or nested to form a position forest structure.

As shown in Figure \ref{fig:PCT}, we illustrate our coding procedure for better understanding. Firstly, we divide the LaTeX sequence into eight substructures: ``$A$'', 
``$y^{3}_{1}$'', ``$+$'', ``$\frac{y^{\beta_{1}}_{2}B}{C}$'',  ``$y^{\beta_{1}}_{2}$'', ``$\beta_{1}$'', ``$B$'' and ``$C$''. The last four substructures are nested within the fraction structure and the others are independent of each other. Secondly, different symbols within a substructure will produce different branches to form a relative position tree.
For instance, in the second substructure ``$y^{3}_{1}$'', the power exponent ``$3$'' and the subscript ``$1$'' of the main body ``$y$'' belong to the upper and lower parts, which are encoded into the left node and right node of the corresponding tree, respectively. 
Once all substructures are encoded into trees, we combine them into a position forest structure. Finally, each symbol is assigned a position identifier in the forest to denote its relative spatial position.

The details of position forest coding are introduced in Algorithm \ref{algorithm0}. A LaTeX sequence $\mathcal{Y}_{c}=\{y_{c}^{(t)}\}_{t=1}^{T}$ is encoded to be an encoded identifier set, $\mathcal{I}=\{\mathbf{I}_{t}\}_{t=1}^{T}$, for representing the position information of all symbols within the sequence. 

\noindent 2) \texttt{Position Forest Decoding:}
The auto-regressive decoding manner has been proven effective for the HMER task\cite{deng2017image,huang2017densely,li2022counting,zhao2022comer,yang2023read}. When decoding the $t$-th symbol, the previously identified $t$-1 symbols are regarded as priors and fed into the decoder. It operates sequentially for $T$ steps, producing a symbol sequence of length $T$. Following this, we employ the type of decoder to parse the positions of all symbols. Differently, the position information of each symbol to be predicted is an identifier (\eg, ``MLLR'') rather than a category. To this end, we divide the position recognition task into two sub-tasks: the nested level prediction task and the relative position prediction task.

First of all, given an expression image and its corresponding identifier set $\mathcal{I}=\{\mathbf{I}_{t}\}_{t=1}^{T}$, we construct the ground truths of the nested level and relative position.
Specifically, for the $k$-th identifier $\mathbf{I}_{k} = \big\{q_{k}^{(i)}| q_{k}^{(i)} \in \{$``M'', ``L'', ``R''$\}\big\}_{i=1}^{\eta_{k}}$, with $\eta_{k}$ as the length of the identifier, we can easily determine that the nested level is $\eta_{k}-1$ and the relative position is $q_{k}^{(\eta_{k})}$. For example, when analyzing the identifier ``MLLR'', we deduce that the symbol resides within a substructure comprising three nested levels, and its relative position is at the lower part (``R'') of the final nested substructure. Accordingly, the ground truth of the nested level is constructed as $\mathcal{Y}_{n}=\big\{y_{n}^{(t)}=\eta_{t}-1|y_{n}^{(t)} \in \{0,1,\cdots,U\}\big\}_{t=1}^{T}$, where $U$ denotes the maximum number of nested levels. The ground truth of the relative position is constructed as $\mathcal{Y}_{r}=\big\{y_{r}^{(t)}=q_{t}^{(\eta_{t})}|y_{r}^{(t)} \in \{$``M'', ``L'', ``R''$\}\big\}_{t=1}^{T}$. 

Subsequently, considering these identifiers $\{\mathbf{I}_{t}\}_{t=1}^{T}$ with varying lengths, we pad them to a uniform length using the ``[pad]'' token. Additionally, we prepend the ``[sos]'' token and append the ``[eos]'' token to each identifier to signify the start and end of the identifier. The processed identifiers are organized into a matrix $\mathbf{Q}=[\mathbf{Q}_{1}, \mathbf{Q}_{2}, \cdots, \mathbf{Q}_{T}]\in \mathbb{R}^{T \times \Upsilon}$, with $\Upsilon$ denotes the uniform length. 

Each vector in the matrix is transcribed into $C$ dimensional identifier embeddings through a nonlinear layer $\xi$, which includes a linear projection, GELU activation, and layer normalization. 
Following the previous transformer-based works\cite{CCD,transformer,zhao2022comer}, a common absolute position encodings $\mathbf{Q}_{\rm pos} \in \mathbb{R}^{T \times C}$ of symbol orders are added to the identifier embeddings.
Thus, the generation of the identifier embedding vector is formulated in the following:
{\setlength\abovedisplayskip{0pt}
\setlength\belowdisplayskip{0pt}
\begin{equation}
\mathbf{Q}_{\rm emb} = [\xi(\mathbf{Q}_{1});\xi(\mathbf{Q}_{2});\cdots;\xi(\mathbf{Q}_{L})] + \mathbf{Q}_{\rm pos}
\end{equation}
}

\noindent where $[\cdot]$ refers to the concatenation operation.

These embedding vectors $\mathbf{Q}_{\rm emb} \in \mathbb{R}^{T \times C}$, along with visual features $\mathbf{V}_{\rm vis} \in \mathbb{R}^{H' \times W' \times C}$ extracted by the backbone, where $\frac{H'}{H}=\frac{W'}{W}=16$, are fed into a three-layer Transformer-based decoder blocks. These blocks primarily consist of Multi-head Attention (MHA), Implicit Attention Correction (IAC, introduced later), and Feed-Forward Network (FFN), processing these inputs to produce output features $\mathbf{F} \in \mathbb{R}^{T \times C}$ for predicting nested levels and relative positions. 

Specifically, at decoding step $t$, $\mathbf{F}_{t} \in \mathbb{R}^{C}$ is taken for predicting the current-step nested level
\begin{equation}
    \left\{
    \begin{aligned}
        p(y_{n}^{(t)}) &= \mathrm{softmax}(\mathbf{W}_{n}\mathbf{F}_{t} + b_{n})\\
    y_{n}^{(t)} &\sim p(y_{n}^{(t)}),
    \end{aligned}
    \right.
\end{equation}

\noindent and the current-step relative position
{\setlength\abovedisplayskip{0pt}
\setlength\belowdisplayskip{0pt}
\begin{equation}
    \left\{
    \begin{aligned}
        p(y_{r}^{(t)}) &= \mathrm{softmax}(\mathbf{W}_{r}\mathbf{F}_{t} + b_{r})\\
    y_{r}^{(t)} &\sim p(y_{r}^{(t)}),
    \end{aligned}
    \right.
\end{equation}
}

\noindent where $\mathbf{W}_{n}$ and $\mathbf{W}_{r}$ are both trainable weights. 

\begin{figure}[t]
  \centering
  \includegraphics[width=0.85\linewidth]{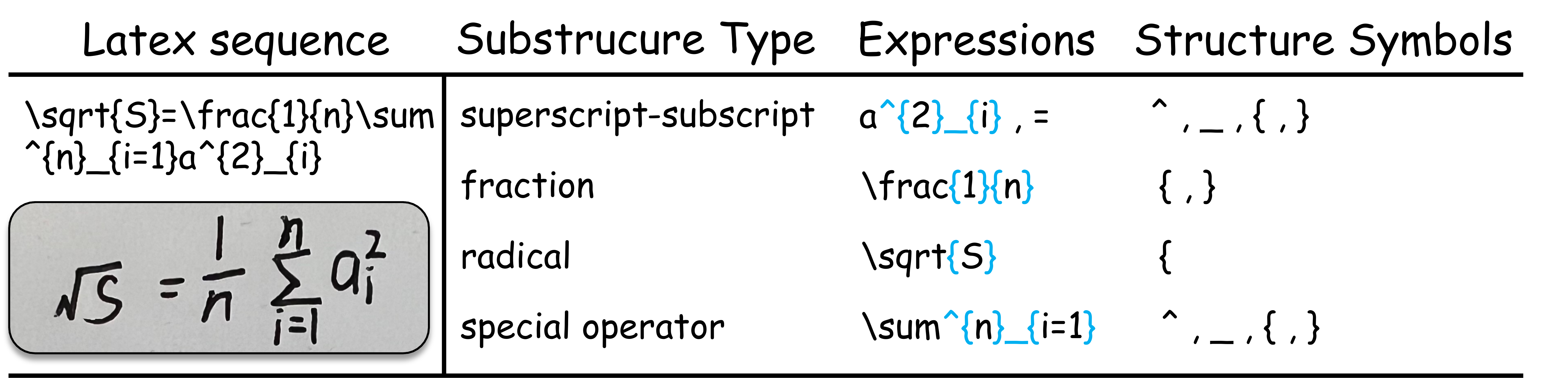}
  \caption{Illustration of structure symbols which are used to describe the position and hierarchical relationships between symbols in LaTeX.}
  \label{fig:substructure}
\end{figure}

\subsubsection{Implicit Attention Correction}\label{IAC}
In HMER, there are over a hundred LaTeX symbols in total, typically divided into two categories:
1) Entity symbols, which have corresponding entities in images;
2) Structure symbols, which have no entity in images and are used to describe the position and hierarchical relationships between entity symbols, as illustrated in Figure \ref{fig:substructure}. When decoding these symbols, coverage problems, \ie, over-parsing and under-parsing, limit the recognition capabilities\cite{zhao2022comer,zhang2018multi,zhang2017watch,li2022counting}.

To address these issues within a transformer structure for HMER, CoMER\cite{zhao2022comer} refines the attention weights of the current decoding step by subtracting the attention of all previous steps for parsing accurately. The corrected attention is then exploited to extract fine-grained feature representations for recognition.

\begin{figure}[t]
  \centering
  \includegraphics[width=1\linewidth]{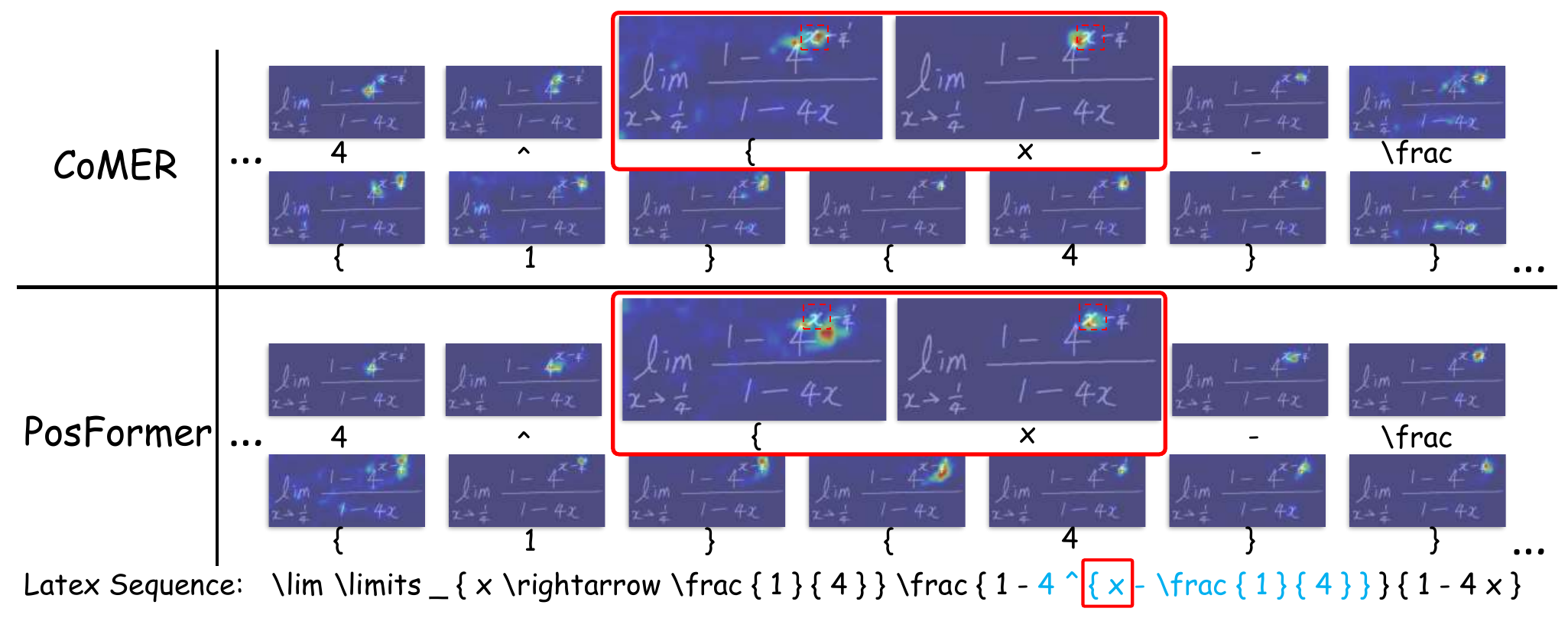}
  \caption{Attention visualization comparisons for the substructure expression ``4\^{}\{x-$\backslash frac\{1\}\{4\}$\}''. CoMER\cite{zhao2022comer} refines the current attention by subtracting all already parsed symbols, leading to alignment-drifted issue. Given that only entity symbols are present on expression images, we introduce zero refinement terms to refine it.}
  \label{fig:attention_issue}
\end{figure}

However, when decoding some structure symbols, we observed that the model allocates more attention to regions that have not yet been parsed or even to the overall image for understanding structure relationships. 
Following the subtraction operation, this mechanism leads to inaccuracies of refined attention in subsequent decoding steps that rely on past alignment information. 
For instance, consider the substructure expression ``4\^{}\{x-$\backslash frac\{1\}\{4\}$\}'' illustrated in Figure~\ref{fig:attention_issue}.
When decoding the structure symbol ``\^{}'' and ``\{', the CoMER model pays more attention to regions that contain ``4'' and ``x'' to understand the superscript relationship. In the subsequent decoding entity symbol step (\eg ``x''), since the previous attention weights are accumulated as a refinement term, the current attention minus the refinement term will lead the model to focus on unimportant regions. 

To this end, we propose a simple and effective correction solution by introducing zero attention as our refinement term. Specifically, when an entity symbol is decoded, we reset the attention weights associated with the preceding structure symbols to zero. 
This is easy to explain: when we encourage the model to generate precise attention for decoding the entity symbol, it suffices to subtract these attention weights from the already parsed entity symbols, given that only entity symbols are present on mathematical expression images.
Therefore, denoting $\mathbf{E}_{k}\in \mathbb{R}^{T \times H' \times W'}$ as the attention weights produced by the $k$-th decoder layer ($k \in \{2,3\}$ in the experiment), 
we propose an indicator function $I_{\Omega}$ to introduce the corresponding refinement term $\mathbf{A}_{k}\in \mathbb{R}^{T \times H' \times W'}$ as follows:
{\setlength\abovedisplayskip{0pt}
\setlength\belowdisplayskip{0pt}
\begin{equation}
I_{\Omega}(y)=\left\{
\begin{array}{ll}
     \mathbf{1} &, \texttt{if} \ y \notin \Omega,  \\
     \mathbf{0} &, \texttt{if} \ y \in \Omega,
\end{array}
\right.
\end{equation}
\begin{equation}
    \mathbf{\Tilde{E}}_{k} = \mathrm{softmax}(\mathbf{E}_{k}),
\end{equation}
\begin{equation}
\mathbf{A}_{k}^{(t)} = \sum_{i=1}^{t-1} \big(\mathbf{\Tilde{E}}_{k}^{(i)} \odot I_{\Omega}(y_{c}^{(i)})\big),
\end{equation}
\begin{equation}
 \mathbf{A}_{k} = [\mathbf{A}_{k}^{(1)}; \mathbf{A}_{k}^{(2)}; \cdots; \mathbf{A}_{k}^{(T)}]
\end{equation}
}

\noindent where $\Omega$ denotes the set of these structure symbols, $\odot$ refers to the Hadamard product, and $[\cdot]$ represents the concatenation operation along the channel.

Subsequently, following the work\cite{zhao2022comer}, the refinement item is introduced to indicate whether an image feature vector has been parsed or not, leading the model to pay more
attention on the unparsed regions. Specifically, the corrected attention weights $\hat{\mathbf{E}}_{k}$ can be calculated using the following formulas:
{\setlength\abovedisplayskip{0pt}
\setlength\belowdisplayskip{0pt}
\begin{equation}
\hat{\mathbf{E}}_{k} =\mathbf{E}_{k}-\phi(\mathbf{A}_{k}),
\end{equation}
}

\noindent where $\phi(\cdot)$ denotes a convolutional layer and a linear layer to extract local coverage information.
Finally, the mechanism achieves a step-wise refinement by collecting past alignment information for each step.

\subsection{Loss Function}
The model is trained end-to-end under a multi-task setting, whose objective is
{\setlength\abovedisplayskip{0pt}
\setlength\belowdisplayskip{0pt}
\begin{equation}
    \texttt{L}_{\rm rec} = -\frac{1}{T}\sum_{t=1}^{T} y_{c}^{(t)} \log p(y_{c}^{(t)}),
\end{equation}
}
{\setlength\abovedisplayskip{0pt}
\setlength\belowdisplayskip{0pt}
\begin{equation}
    \texttt{L}_{\rm pos} = -\frac{1}{T}\sum_{t=1}^{T} \big(y_{n}^{(t)} \log p(y_{n}^{(t)}) + y_{r}^{(t)} \log p(y_{r}^{(t)})\big),
\end{equation}
}

\noindent where $\mathcal{Y}_{c}=\{y_{c}^{(t)}\}_{t=1}^{T}$ denotes the groundtruth LaTeX sequence, $\mathcal{Y}_{n}=\{y_{n}^{(t)}\}_{t=1}^{T}$ is the groundtruth nested level, $\mathcal{Y}_{r}=\{y_{r}^{(t)}\}_{t=1}^{T}$ refers to the groundtruth relative position. And $p(y_{c}^{(t)})$, $p(y_{n}^{(t)})$, and $p(y_{r}^{(t)})$ denotes the predicted distributions of three tasks.
Finally, the overall training loss is summarized as:
{\setlength\abovedisplayskip{0pt}
\setlength\belowdisplayskip{0pt}
\begin{equation}
  \label{eq:3}
    \begin{aligned}
        \texttt{L}_{\rm all} = \lambda_{1} \cdot  \texttt{L}_{\rm rec}  + \lambda_{2} \cdot  \texttt{L}_{\rm pos},
    \end{aligned}
  \end{equation}
}

\noindent where $\lambda_{1}$ and $\lambda_{2}$ are a loss coefficient, set as 1 by default.

\section{Experiment}

\subsection{Datasets}\label{dataset}
\textbf{CROHME}\cite{mouchere2014icfhr,mouchere2016icfhr2016,mahdavi2019icdar} is a publicly available single-line handwritten mathematical expression benchmark. The training set consists of 8836 expression images. The CROHME 2014\cite{mouchere2014icfhr}/2016\cite{mouchere2016icfhr2016}/2019\cite{mahdavi2019icdar} test sets include 986, 1147, and 1199 expression images, respectively. 

\noindent \textbf{M$^{2}$E}\cite{yang2023read} is a multi-line handwritten mathematical expression benchmark, which is collected from real-world scenarios, including math papers, exercise books, handwriting works, \etc. The dataset contains 79,979 training images, 9,992 validating images, and 9,985 testing images.

\noindent \textbf{MNE} We construct a \textbf{M}ulti-level \textbf{N}ested handwritten mathematical \textbf{E}xpression test set, used to evaluate a model's ability to recognize complex expression images. See the supplementary materials for more details.

\subsection{Metrics}
ExpRate, $\leq1$, $\leq2$, and $\leq3$ metrics are widely used to measure the performance in HMER. These metrics represent the expression recognition rate when we tolerate 0 to 3 symbol-level errors.
We also follow \cite{yang2023read} and adopt the Character Error Rate (CER) to evaluate the performance on the M$^{2}$E dataset.

\subsection{Implementation Details}
Following the CoMER\cite{zhao2022comer} method, we employ a DenseNet architecture as the backbone, a three-layer decoder as our position forest decoder, and a linear layer for the symbol classification. For the position recognition task, two linear layers with different output dimensions are employed, as the maximum number of nested levels is 3 and the vocabulary size of the relative position is 6, including ``[sos]'', ``[eos]'', ``[pad]'', ``M'', ``L'', and ``R''. For multi-line formulas, each line is first converted into several trees corresponding to the substructures. Thanks to our forest structure, the trees corresponding to different lines can be easily arranged in series to form a forest.
Furthermore, we also follow the same training parameters of CoMER\cite{zhao2022comer} and LAST\cite{yang2023read}, including batch size, learning rate, optimizer, and training epochs for fair comparisons on single-line CROHME-series datasets and multi-line M$^{2}$E dataset, respectively. All experiments are executed on a server equipped with a single NVIDIA A800 GPU.
\newcolumntype{P}[1]{>{\centering\arraybackslash}p{#1}}
\newcolumntype{L}[1]{>{\arraybackslash}p{#1}}
\begin{table}[!t]
    \centering
    \caption{Performance comparison with previous SOTA methods on CROHME 2014/2016/2019 test sets (in \%). ``Scale-aug'' refers to the scale augmentation\cite{li2020improving}.}
    \scalebox{0.9}{
    \begin{tabular}{c|c|P{3cm}|L{2cm}|P{1cm}P{1cm}P{1cm}}
    \toprule 
    \textbf{Dataset} & \textbf{Scale-aug} & \textbf{Model} & \textbf{ExpRate} $\uparrow$ & $\bf{\leq1}$ $\uparrow$ & $\bf{\leq2}$ $\uparrow$ & $\bf{\leq3}$ $\uparrow$ \\
    \midrule
    \multirow{10}{*}{ CROHME 14} &\multirow{6}{*}{$\times$} & DWAP\cite{zhang2018multi}  & 50.10 & - & - & - \\
    \multirow{10}{*}{}&\multirow{6}{*}{} & BTTR\cite{zhao2021handwritten}& 53.96 & 66.02 & 70.28 & - \\
    \multirow{10}{*}{}&\multirow{6}{*}{} & TSDNet\cite{TSDNet}& 54.70  & 68.85 & 74.48 & - \\ 
    \multirow{10}{*}{}&\multirow{6}{*}{} & SAN\cite{yuan2022syntax} & 56.2  & 72.6 & 79.2 & - \\
    \multirow{10}{*}{}&\multirow{6}{*}{} & CAN\cite{li2022counting} & 57.26  & 74.52 & 82.03 & - \\ 
    \multirow{10}{*}{}&\multirow{6}{*}{} & PosFormer (ours) & $\mathbf{60.45}_{(+3.19)}$ & \textbf{77.28} & \textbf{83.68} & \textbf{87.83} \\
    \cline{2-7}
    \multirow{10}{*}{}&\multirow{4}{*}{\checkmark} & Li \emph{et al.}\cite{li2020improving}  & 56.59 & 69.07 & 75.25 & 78.60 \\
    \multirow{10}{*}{}&\multirow{4}{*}{}& CoMER\cite{zhao2022comer} & 59.33 & 71.70 & 75.66 & 77.89 \\
    \multirow{10}{*}{}&\multirow{4}{*}{}& BPD-Coverage\cite{li2024tree}& 60.65 &- & - & - \\
    \multirow{10}{*}{}&\multirow{4}{*}{}& PosFormer (ours) &$\mathbf{62.68}_{(+2.03)}$  &\textbf{79.01} &\textbf{84.69}  &\textbf{88.84} \\
    \midrule
    \multirow{10}{*}{ CROHME 16} &\multirow{6}{*}{$\times$}& DWAP\cite{zhang2018multi}  & 47.50 & - & - & - \\
    \multirow{10}{*}{} & \multirow{6}{*}{}& BTTR\cite{zhao2021handwritten}  & 52.31 & 63.90 & 68.61 & - \\
    \multirow{10}{*}{} &\multirow{6}{*}{} & TSDNet\cite{TSDNet} & 52.48  & 68.26 & 73.41 & - \\
    \multirow{10}{*}{} &\multirow{6}{*}{} &SAN\cite{yuan2022syntax} & 53.6  & 69.6 & 76.8 & - \\
    \multirow{10}{*}{} & \multirow{6}{*}{}& CAN\cite{li2022counting} & 56.15 & 72.71 & 80.30 & - \\ 
    \multirow{10}{*}{} & \multirow{6}{*}{}& PosFormer (ours) &$\mathbf{60.94} _{(+4.79)}$&\textbf{76.72} &\textbf{83.87} &\textbf{88.06} \\
    \cline{2-7}
    \multirow{10}{*}{} & \multirow{4}{*}{\checkmark}& Li \emph{et al.}\cite{li2020improving}  & 54.58 & 69.31 & 73.76 & 76.02 \\
    \multirow{10}{*}{}&\multirow{4}{*}{}& BPD-Coverage\cite{li2024tree} & 58.50 &- & - & - \\
    \multirow{10}{*}{} & \multirow{4}{*}{}& CoMER\cite{zhao2022comer} & 59.81 & 74.37 & 80.30 & 82.56 \\
    \multirow{10}{*}{} & \multirow{4}{*}{}& PosFormer (ours)&$\mathbf{61.03}_{(+1.22)} $&\textbf{77.86} &\textbf{84.74} &\textbf{89.28} \\
    \midrule
    \multirow{9}{*}{CROHME 19}&\multirow{5}{*}{$\times$}& BTTR \cite{zhao2021handwritten} & 52.96 & 65.97 & 69.14 & - \\
    \multirow{9}{*}{} &\multirow{5}{*}{} &SAN\cite{yuan2022syntax} & 53.5& 69.3 & 70.1 & - \\
    \multirow{9}{*}{} &\multirow{5}{*}{}& CAN\cite{li2022counting} & 55.96 & 72.73 & 80.57 & - \\
    \multirow{9}{*}{} &\multirow{5}{*}{} &TSDNet\cite{TSDNet}& 56.34  & 72.97 & 77.84 & - \\
    \multirow{9}{*}{} &\multirow{5}{*}{}& PosFormer (ours)&$\mathbf{62.22}_{(+5.88)}$ &\textbf{79.40} &\textbf{86.57} &\textbf{89.99} \\
    \cline{2-7}
    \multirow{9}{*}{} & \multirow{4}{*}{\checkmark}& Ding \emph{et al.}\cite{Ding2021encoder}& 61.38 & 75.15 & 80.23 & 82.65 \\
    \multirow{9}{*}{}&\multirow{4}{*}{}& BPD-Coverage\cite{li2024tree}& 61.47&- & - & - \\
    \multirow{9}{*}{} & \multirow{4}{*}{}& CoMER\cite{zhao2022comer} & 62.97 & 77.40 & 81.40 & 83.07 \\
    \multirow{9}{*}{} & \multirow{4}{*}{}& PosFormer (ours)&$\mathbf{64.97} _{(+2.00)}$&\textbf{82.49}  &\textbf{87.24} &\textbf{90.16} \\
    \bottomrule
    \end{tabular}
    }
    \label{CROHME}
\end{table}
\begin{table}[t]
    \centering
    \caption{Comparisons with previous SOTA methods on the HME100k dataset.}
    \scalebox{0.9}{
    \begin{tabular}{l|L{1.9cm}|P{1cm}P{1cm}P{1cm}}
    \toprule 
    \textbf{Model} & \textbf{ExpRate} $\uparrow$ & $\bf{\leq1}$ $\uparrow$ & $\bf{\leq2}$ $\uparrow$ & $\bf{\leq3}$ $\uparrow$  \\
    \midrule 
    SAN~\cite{yuan2022syntax}  & 67.10 & - &- &- \\
    CAN~\cite{li2022counting}  & 67.31 & 82.93 &  89.17& -  \\ 
    \midrule
    PosFormer (ours) & $\mathbf{69.51}_{(+2.20)}$ & \textbf{84.91} & \textbf{90.51} & \textbf{93.25} \\
    \bottomrule
    \end{tabular}
    }
    \label{HME100k}
\end{table}
\begin{table}[t]
    \centering
    \caption{Performance comparison with previous SOTA methods on the multi-line $\mathrm{M}^2 \mathrm{E}$ dataset. ExpRate, $\leq1, \leq2, \leq3$ are shown in percentage (\%).}
    \scalebox{0.9}{
    \begin{tabular}{l|L{1.9cm}|P{1cm}P{1cm}P{1cm}|P{1.2cm}}
    \toprule 
    \textbf{Model} & \textbf{ExpRate} $\uparrow$ & $\bf{\leq1}$ $\uparrow$ & $\bf{\leq2}$ $\uparrow$ & $\bf{\leq3}$ $\uparrow$ &\textbf{CER} $\downarrow$ \\
    \midrule 
    DWAP\cite{zhang2018multi}  & 53.14 & 70.15 & 78.34 & 82.69 & 0.0517 \\ 
    Vanilla Transformer \cite{transformer}  & 55.15 & 71.79 & 79.46 & 83.39 & 0.0576 \\
    WS-WAP\cite{truong2020improvement} & 55.20 & 72.13 & 80.13 & 83.98 & 0.0555 \\
    BTTR\cite{zhao2021handwritten}  & 55.25 & 72.09 & 80.17 & 84.33 & 0.0528 \\
    ABM\cite{bian2022handwritten} & 55.48 & 72.05 & 80.06 & 83.96 & 0.0552 \\
    CoMER\cite{zhao2022comer} & 56.20 & 73.39 & 81.11 & 84.94 & 0.0499 \\
    LAST\cite{yang2023read}  & 56.50 & 72.80 & 80.81 & 84.61 & 0.0530\\
    \midrule
    PosFormer (ours) & $\mathbf{58.33}_{(+1.83)}$ & \textbf{75.58} & \textbf{83.22} & \textbf{86.86} &\textbf{0.0366}\\
    \bottomrule
    \end{tabular}
    }
    \label{M2E}
\end{table}
\begin{table}[!t]
  \centering
  \caption{Evaluate the effectiveness of the proposed modules on the CROHME-series datasets. ``PF'' and `IAC'' denote the Position Forest and Implicit Attention Correction module, respectively. ExpRate, $\leq1, \leq2$ are shown in percentage (\%).}
  \scalebox{0.75}{
    \begin{tabular}{l|L{2cm}P{1cm}P{1cm}|L{2cm}P{1cm}P{1cm}|L{2cm}P{1cm}P{1cm}}
    \toprule
    \multirow{2}{*}{ Model } & \multicolumn{3}{c|}{ CROHME 2014} & \multicolumn{3}{c|}{ CROHME 2016 } & \multicolumn{3}{c}{ CROHME 2019 } \\
    \cline { 2 - 10 } & \textbf{ExpRate} $\uparrow$ & $\bf{\leq1}$ $\uparrow$ & $\bf{\leq2}$ $\uparrow$ & \textbf{ExpRate} $\uparrow$ & $\bf{\leq1}$ $\uparrow$ & $\bf{\leq2}$ $\uparrow$ & \textbf{ExpRate} $\uparrow$ & $\bf{\leq1}$ $\uparrow$ & $\bf{\leq2}$ $\uparrow$ \\
    \hline baseline & 59.33&71.70 &75.66 &59.81 & 74.37&80.30 &62.97 &77.40 &81.40 \\
    + PF & $62.13_{(+2.80)}$&78.88& 84.77& $61.03_{(+1.22)}$& 77.59& \textbf{85.35} &$63.80_{(+0.83)}$&  80.90 &86.49  \\
    + IAC & $\mathbf{62.64}_{(+3.31)}$& \textbf{79.29}&\textbf{85.08} &$\mathbf{61.20}_{(+1.39)}$ &\textbf{78.29} &84.83&$\mathbf{64.64}_{(+1.67)}$ & \textbf{82.40}&\textbf{87.16} \\
    \bottomrule
    \end{tabular}
}
\label{tb:Model Structure}
\end{table}

\subsection{Comparison with State-of-the-Art Methods}
\noindent \textbf{Results on Single-line Datasets}
First, we exhibit comparison results between PosFormer and previous SOTA methods on the CROHME datasets in Table \ref{CROHME}. Specifically, we provide performance results of PosFormer with and without scale augmentation for a fair comparison. Without scale augmentation, PosFormer surpasses the previous SOTA results by 3.19\%, 4.79\% and 5.88\% on the CROHME 14/16/19 test set, respectively. When using the scale augmentation, PosFormer further refreshes the recognition results, achieving gains of 2.03\%, 1.22\%, and 2.00\% in the ExpRate metric.
Second, we also conduct the comparative experiments on large-scale HME100k dataset in Table \ref{HME100k} and PosFormer significantly exceeds previous works. 
Finally, PosFormer exhibits a significant performance improvement in terms of symbol-level error tolerance metrics, which demonstrates the correction ability of PosFormer when recognizing complex expressions.

\noindent \textbf{Results on Multi-line Dataset}
Compared to the single-line CROHME-series datasets, the multi-line handwritten mathematical expression dataset, M2E, contains a larger number of images with complex structures and long sequences. To demonstrate the effectiveness and robustness of our model, we compare PosFormer with the previous SOTA methods on the M2E dataset in Table \ref{M2E}. Specifically, PosFormer achieves the highest performance, with an ExpRate metric of 58.33\% and a CER metric of 0.0366. This represents a 2.13\% and 1.83\% improvement over the CoMER method and the latest LAST\cite{yang2023read} method, respectively. The latter is specifically tailored for decoding multi-line expressions. 

\section{Ablations and Analysis}

\noindent \textbf{Effectiveness of Network Components.}
We demonstrate the performance gains brought by the two components of PosFormer, Position Forest (PF) and Implicit Attention Correction (IAC), respectively, as shown in Table \ref{tb:Model Structure}.
When the baseline model introduces PF to assist expression recognition by explicitly parsing the relative position relationships of symbols in mathematical formulas, its performance improves by 2.80\%, 1.22\%, and 0.83\% on the CROHME 14/16/19 test sets, respectively. 
This improvement highlights that taking position recognition as an auxiliary task can effectively improve the recognition ability of sequence-based methods.
Additionally, the performance gains significantly improve when allowing for 1 and 2 errors, indicating that the PF module effectively boosts the model's correction capacity to recognize complex mathematical formulas.
Building upon this, integrating the IAC module further enhances performance on the test sets, with increases of 0.51\%, 0.17\%, and 0.84\% on the CROHME 14/16/19 test sets, respectively.

\noindent \textbf{Extensibility to RNN-based methods.}
Position Forest (PF) can easily be encapsulated into a plug-in component. To demonstrate its robustness and generalizability, we extend it to RNN-based methods, as shown in Table \ref{tab:plug}.
Specifically, we selected three mainstream methods\cite{li2022counting,zhang2018multi,bian2022handwritten} as baseline models to conduct comparative experiments, and used the same decoder as these methods to parse positions.
As a result, ``DWAP+PF'' exhibits performance enhancements of 7.00\%/8.73\% on the CROHME 14/16 test sets, respectively. ``ABM+PF'' improves model accuracy by 1.26\%/3.58\%/0.34\%, respectively. Similarly, with CAN as the baseline, ``CAN+PF'' increases recognition performance by 2.03\%/ 2.87\%/4.17\%, respectively. This implies that PF can be generalized to existing RNN-based models to significantly boost their performance.
\begin{table}[t]
  \centering
  \caption{Performance gains on CROHME-series datasets~\cite{mouchere2014icfhr,mouchere2016icfhr2016,mahdavi2019icdar} when extended our proposed Position Forest (PF) to previous RNN-based methods. $\dagger$ indicates our reproduced result. ExpRate, $\leq1, \leq2$ are shown in percentage (\%).}
  \scalebox{0.75}{
    \begin{tabular}{l|L{2cm}P{1cm}P{1cm}|L{2cm}P{1cm}P{1cm}|L{2cm}P{1cm}P{1cm}}
    \toprule
    \multirow{2}{*}{ Model } & \multicolumn{3}{c|}{ CROHME 2014} & \multicolumn{3}{c|}{ CROHME 2016 } & \multicolumn{3}{c}{ CROHME 2019 } \\
    \cline { 2 - 10 } & \textbf{ExpRate} $\uparrow$ & $\bf{\leq1}$ $\uparrow$ & $\bf{\leq2}$ $\uparrow$ & \textbf{ExpRate} $\uparrow$ & $\bf{\leq1}$ $\uparrow$ & $\bf{\leq2}$ $\uparrow$ & \textbf{ExpRate} $\uparrow$ & $\bf{\leq1}$ $\uparrow$ & $\bf{\leq2}$ $\uparrow$ \\
    \hline DWAP\cite{zhang2018multi} & 50.10& -&- &47.50&- &- & -&- &- \\
    DWAP+PF (ours) & $\mathbf{57.10}_{(+7.00)}$&\textbf{73.02}& \textbf{80.53}& $\mathbf{56.23}_{(+8.73)}$& \textbf{72.24}& \textbf{81.17} &$\mathbf{57.30}$&  \textbf{75.56} &\textbf{82.24}  \\
    \hline ABM\cite{bian2022handwritten} & 56.85& 73.73 &81.24 &52.92& 69.66 &78.73& 53.96 &71.06& 78.65   \\
     ABM+PF (ours) & $\mathbf{58.11}_{(+1.26)}$&\textbf{74.54} & \textbf{80.02}&$\mathbf{56.50}_{(+3.58)}$ &\textbf{72.89} &\textbf{80.47} &$\mathbf{54.30}_{(+0.34)}$ &\textbf{73.90} &\textbf{82.90} \\
    \hline CAN$^{\dagger}$\cite{li2022counting} & 55.27&72.41& 80.43& 53.97& 70.27& 78.03 &52.96&  72.31 &79.82  \\
    CAN+PF (ours) & $\mathbf{57.30}_{(+2.03)}$& \textbf{73.94}&\textbf{80.12} &$\mathbf{56.84}_{(+2.87)} $&\textbf{73.76} & \textbf{82.30}&$\mathbf{57.13}_{(+4.17)}$ & \textbf{74.73}&\textbf{82.07} \\
    \bottomrule
    \end{tabular}
}
\label{tab:plug}
\end{table}

\noindent \textbf{Other comparisons.} 
1) Position-aware model: we add a comparison with positional enhancement work~\cite{yue2020robustscanner} on the CROHME dataset in Table \ref{tab:6}, and the average gain is 3.69\%.
2) Language-aware model: introducing language models is a promising direction to further improve performance. The visual outputs of some HMER methods (\emph{e.g.}, RLFN~\cite{chen2023language}) are fed into a language model~\cite{liu2019roberta} to implement recognition correction with linguistic context. Although PosFormer is a language-free method, it still gets a 6.25\% gain as shown in Table \ref{tab:5}. 
\begin{table}[t]
\centering 
\caption{Average ExpRate results on CHROME 14/16/19.} 
\begin{minipage}{.5\linewidth}
\centering
\scalebox{0.8}{
\begin{tabular}{c|c|c|c}
\toprule
Model & DWAP & DWAP+RobustScanner & DWAP+PF \\
\midrule
\textbf{ExpRate} $\uparrow$ & 49.17 & 53.18 & \textbf{56.87}$_{(+3.69)}$ \\
\bottomrule
\end{tabular}
}
\label{tab:5}
\end{minipage}%
\hfill
\begin{minipage}{.45\linewidth}
\centering
\scalebox{0.8}{
\begin{tabular}{c|c|c}
\toprule
Model & RLFN & PosFormer \\
\midrule
\textbf{ExpRate} $\uparrow$ & 55.01 & \textbf{61.26}$_{(+6.25)}$ \\
\bottomrule
\end{tabular}
}
\label{tab:6}
\end{minipage}
\end{table}

\section{Conclusion}
We propose an effective position forest transformer (PosFormer) for handwritten mathematical expression recognition (HMER) by adding a component of positional understanding to sequence-based methods. 
For each mathematical expression, we first encode it as a forest structure without requiring extra annotations, and then parse their nested levels and relative positions in the forest. By optimizing the position recognition task to assist HMER, PosFormer explicitly enables position-aware symbol-level feature representation learning in complex and nested mathematical expressions. Extensive experiments validate the performance superiority of PosFormer without introducing additional latency or computational cost during inference. This highlights the significance of explicitly modelling the position relationship of expressions in sequence-based methods.

\noindent \textbf{Acknowledgment.} 
This work was supported by the NSFC under Grant 62176159 and 62322604, and in part by the Shanghai Municipal Science and Technology Major Project under Grant 2021SHZDZX0102.

\par\vfill\par

\newpage
\appendix

\definecolor{lightcyan}{rgb}{0.78, 0.9, 0.9}
\sethlcolor{lightcyan} 
\definecolor{cvprblue}{rgb}{0.21,0.49,0.74}

\title{PosFormer: Recognizing Complex\\ Handwritten Mathematical Expression with Position Forest Transformer \\(Supplementary Material)} 

\titlerunning{Position Forest Transformer for HMER}

\author{Tongkun Guan\inst{1}\textsuperscript{$*$}\orcidlink{0000-0003-3346-8315} \and
Chengyu Lin\inst{2}\textsuperscript{$*$}\orcidlink{0009-0000-2740-2315} \and
Wei Shen\inst{1(\textrm{\Letter})} \and
Xiaokang Yang\inst{1}
}

\authorrunning{T.~Guan et al.}

\institute{MoE Key Lab of Artificial Intelligence, AI Institute, Shanghai Jiao Tong University \email{\{gtk0615,wei.shen\}@sjtu.edu.cn} \and
Paris Elite Institute of Technology, Shanghai Jiao Tong University
\email{lacayqwq@sjtu.edu.cn}
}

\maketitle

\section{MNE}
\label{sec:intro}

\subsection{Structural Complexity Definition} 
The structural complexity of a ME is defined as the maximum nested levels of its substructures (\emph{e.g., the nested level of $x^{2^{2}} + x^{2^{2^{2}}}+ x^{2^{2^{2_{2}}}}$ is 4}). 
MEs with nested levels above 2 are considered complex. 

\subsection{Data construct} 
 We construct a \textbf{M}ulti-level \textbf{N}ested handwritten mathematical \textbf{E}xpression test set, used to evaluate a model's ability to recognize complex expression images. MNE includes three subsets (N1, N2, and N3) with nested levels of 1, 2, and 3, respectively. N4 is ignored because it accounts for only 0.2\% of all public datasets, while the other levels are 37.4\%, 51.4\%, 9.7\%, and 1.3\%, respectively (statistics). The subsets N1, N2 from CROHME, and N3 were collected by us.
 
 Specifically, we first collect these subsets from the CHOHME test sets, according to the number of nested levels of expressions, including 1875, 304, and 10 images. Subsequently, to more convincingly assess the model’s ability to identify complex mathematical expressions,  we further expand the sample number of subset N3 to 1464 images by collecting complex expression images from public documents~\cite{wang2024unimernet,mouchere2014icfhr,mouchere2016icfhr2016,yuan2022syntax} and real-world handwriting homework. 

\newcolumntype{P}[1]{>{\centering\arraybackslash}p{#1}}
\newcolumntype{L}[1]{>{\arraybackslash}p{#1}}
\begin{table}[t]
    \centering
    \caption{Performance comparison with previous SOTA methods on the complex MNE test set. ExpRate, $\leq1, \leq2, \leq3$ are shown in percentage (\%).}
    \scalebox{0.9}{
    \begin{tabular}{c|P{1.0cm}|P{3cm}|L{2.0cm}|P{1.2cm}P{1.2cm}P{1.2cm}}
    \toprule 
    \textbf{Dataset} &\textbf{Size}& \textbf{Model} & \textbf{ExpRate} $\uparrow$ & $\bf{\leq1}$ $\uparrow$ & $\bf{\leq2}$ $\uparrow$ & $\bf{\leq3}$ $\uparrow$ \\
    \midrule 
    \multirow{2}{*}{N1} &\multirow{2}{*}{1875} &COMER   &59.73 &77.55 &84.11 &88.91\\
    & &PosFormer (ours)   &$\mathbf{60.59}_{(+0.86)}$ &\textbf{77.97} &\textbf{84.32} &\textbf{88.75}\\
    \midrule
    \multirow{2}{*}{N2} &\multirow{2}{*}{304} &COMER   &37.17 &53.95 &65.13 &72.37\\
    & &PosFormer (ours)   &$\mathbf{38.82}_{(+1.65)}$ &\textbf{56.91} &\textbf{66.12} &\textbf{73.36}\\
    \midrule
    \multirow{4}{*}{N3} &\multirow{4}{*}{1464} &SAN   &8.61 & 13.18&16.46 &20.82\\
    & &CAN   &7.72 & 10.52&12.43 &15.44\\
    & &COMER   &24.04 & 32.31&36.34 &39.89\\
    & &PosFormer (ours)   &$\mathbf{34.08}_{(+10.04)}$ &\textbf{36.82} &\textbf{40.30} &\textbf{43.10}\\
    \bottomrule
    \end{tabular}
    }
    \label{MNE}
\end{table}
\subsection{Results on MNE Datasets} As shown in Table \ref{MNE}, we conduct the comparison experiments with the previous SOTA methods on these subsets. The results, derived from loading the optimal checkpoint for single-line dataset performance evaluation, show our method achieving performance gains of 0.86\%, 1.65\%, and 10.04\% on the three subsets, respectively. Moreover, as the complexity of the subsets increases, the performance improvements afforded by PosFormer further escalate, which underscores the significance of enhancing position-aware symbol feature extraction.

\subsection{Visualization.}
\begin{figure}[h]
  \centering
  \includegraphics[width=\linewidth]{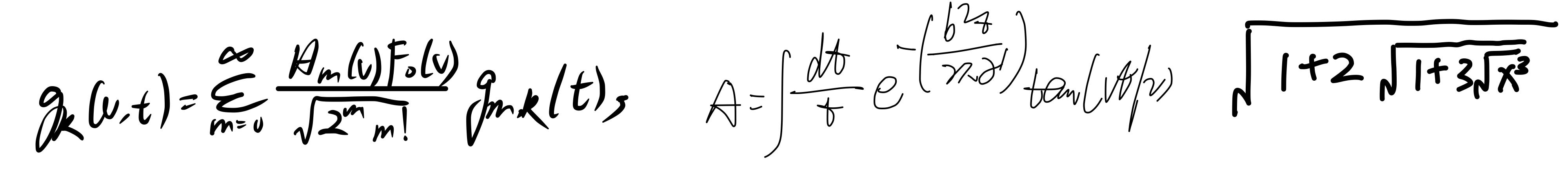}
  \caption{Some visualization examples.}
  \label{fig:complex_case}
\end{figure}

\section{Model Cost} 
We add speed and parameter comparisons in Table \ref{tab:fps}. The PF (69K, removed during inference) and IAC (0K) are introduced to \textbf{baseline}. We add the comparisons of parameters and latency during inference on the same platform.
\begin{table}[t]
\centering
\caption{Comparison results of speed and parameter amount.}
\scalebox{0.9}{
\begin{tabular}{P{2cm}|P{1.5cm}|P{2cm}|P{1.5cm}|P{2cm}|P{2.5cm}}
\toprule
\textbf{Model} &CoMER & PosFormer & DWAP & DWAP+PF & DWAP+CAN \\
\midrule
\textbf{\#Params}&6.4M & 6.4M & 4.7M & 4.7M & 17.0M \\
\textbf{FPS} ($\uparrow$)& 6.30 & 6.28 & 21.73 & 21.72 & 18.47  \\
\bottomrule
\end{tabular}
}
\label{tab:fps}
\end{table}

\begin{table}[t]
    \centering
    \caption{Comparison with tree-based methods on CROHME-series datasets. ``PF'' denotes the Position Forest. ExpRate, $\leq1, \leq2$ are shown in percentage (\%).}
    \scalebox{0.75}{
        \begin{tabular}{l|L{2cm}P{1cm}P{1cm}|L{2cm}P{1cm}P{1cm}|L{2cm}P{1cm}P{1cm}}
    \toprule
    \multirow{2}{*}{ Model } & \multicolumn{3}{c|}{CROHME 2014} & \multicolumn{3}{c|}{ CROHME 2016 } & \multicolumn{3}{c}{CROHME 2019 } \\
    \cline { 2 - 10 } & \textbf{ExpRate} $\uparrow$ & $\bf{\leq1}$ $\uparrow$ & $\bf{\leq2}$ $\uparrow$ & \textbf{ExpRate} $\uparrow$ & $\bf{\leq1}$ $\uparrow$ & $\bf{\leq2}$ $\uparrow$ & \textbf{ExpRate} $\uparrow$ & $\bf{\leq1}$ $\uparrow$ & $\bf{\leq2}$ $\uparrow$ \\
    \hline 
    DWAP & 50.10& -&- &47.50&- &- & -&- &- \\
    DWAP+SAN & 50.41& 68.15&76.06 &51.87&68.70 &75.68 &51.13&71.23 &79.23 \\
    \hline
    DWAP+PF (ours) & $\textbf{57.10}_{(+6.69)}$&\textbf{73.02}& \textbf{80.53}& $\mathbf{56.23}_{(+4.36)}$& \textbf{72.24}& \textbf{81.17} &$\mathbf{57.30}_{(+6.17)}$&  \textbf{75.56} &\textbf{82.24}  \\
    \bottomrule
    \end{tabular}
    }
    \label{tab:tree}
\end{table}
\section{Difference to Tree-based Methods}
\noindent \textbf{Different purposes.} 
Tree-based methods are designed for HMER directly. Position Forest (PF) assists sequence-based methods by explicitly modelling positional relationships between symbols, which is removed during inference and incurs no additional latency or computational cost. 

\noindent \textbf{Different mechanisms.} 
Tree-based methods model a mathematical expression as a syntax tree, achieving expression recognition by predicting the entire tree and assembling these entity symbols (nodes) and syntactic relationships (edges). PF models the mathematical expression as a position forest structure, reflecting the actual spatial positions of symbols within the LaTeX expression's substructures. The simple nested levels and relative positions are parsed to positively promote position-aware symbol-level feature extraction. Accordingly, when separately integrating them into sequence-based methods to enhance structural relationship perception, PF can simultaneously predict the category and position for each symbol of the sequence in the same decoder, while the target context of the extra tree-based decoder conflicts with the sequence. We also conduct the ablation experiment in Table \ref{tab:tree}, comparing with the latest open-source tree-based method SAN, PosFormer brings significant gains.

\noindent \textbf{Explanation Example.}
Tree-based methods encode an expression as a syntax tree based on complete triplet relationships (parent node, child node, parent-child relations). The absence of any of these relationships will cause encoding failure; \emph{e.g.,} $\mathrm{^{14}CO_{2}}$, as there is no parent node for ``$1$'' (\textcolor{cvprblue}{\texttt{\^{}\{14\}CO\_\{2\}}}). 
Differently, without strict syntax dependency, the \textbf{key} (``\texttt{\^{}}'') triggers our PF to view ``1'' as the upper part in the ME image, thereby encoding the relative position and nested level (we need) is ``upper (L)'' and 1 to assist training, not depending on other symbols.
Moreover, compared to the limitations associated with tree-based methods, PF employs the sequence-based decoding process, where each ME case can be parsed into a sequence. 

In summary, usage differences between tree-based methods (A) and our method (B) lie in: 

\noindent \textbf{1) Simplify encoding.} 

(A) ME\raisebox{-0.2ex}[0pt][0pt]{$\textcolor{cvprblue}{\xrightarrow{\smash{\raisebox{-0.2ex}{\text{\emph{syntax rules}}}}}}$}a syntax tree\raisebox{-0.2ex}[0pt][0pt]{$\textcolor{cvprblue}{\xrightarrow{\smash{\raisebox{-0.2ex}{\text{\emph{target}}}}}}$}complete triple tuple (parent, child, seven types of parent-child relationships). 

(B) ME\raisebox{-0.2ex}[0pt][0pt]{$\textcolor{cvprblue}{\xrightarrow{\smash{\raisebox{-0.2ex}{\text{\emph{split}}}}}}$}substructures\raisebox{-0.3ex}[0pt][0pt]{$\textcolor{cvprblue}{\xrightarrow{\smash{\raisebox{-0.2ex}{\small\text{\emph{M/L/R}}}}}}$}independent spatial position trees
\raisebox{-0.2ex}[0pt][0pt]{$\textcolor{cvprblue}{\xrightarrow{\smash{\raisebox{-0.2ex}{\text{\emph{form}}}}}}$}

\noindent forest\raisebox{-0.2ex}[0pt][0pt]{$\textcolor{cvprblue}{\xrightarrow{\smash{\raisebox{-0.2ex}{\text{\emph{target}}}}}}$}nested levels, relative positions.

\noindent \textbf{2) Assist decoding.} T/I denotes a training/inference stage.

(A) Image\raisebox{-0.2ex}[0pt][0pt]{$\textcolor{cvprblue}{\xrightarrow{\smash{\raisebox{-0.2ex}{\text{\emph{stage1}}}}}}$}triple tuple\raisebox{-0.2ex}[0pt][0pt]{$\textcolor{cvprblue}{\xrightarrow{\smash{\raisebox{-0.2ex}{\text{\emph{stage2}}}}}}$}LaTeX sequence (T/I). 

(B) Image\raisebox{-0.3ex}[0pt][0pt]{$\textcolor{cvprblue}{\xrightarrow{\smash{\raisebox{-0.2ex}{\text{\emph{one stage}}}}}}$}LaTeX sequence (T/I) + nested levels(\textbf{T}) + relative position(\textbf{T}), as PF explicitly promotes position-aware symbol-level representation learning at the feature level during training.
%
%
\bibliographystyle{splncs04}
\bibliography{main}
\end{document}